\relax
\documentclass[letterpaper]{article} 
\usepackage{helvet}  
\usepackage{courier}  
\usepackage{url}  
\usepackage{graphicx}  

\usepackage[namelimits]{amsmath} 
\usepackage{amssymb}             
\usepackage{amsfonts}            
\usepackage{mathrsfs}            
\usepackage{algorithm}
\usepackage{algorithmic}
\usepackage{makecell,rotating,multirow,diagbox}
\usepackage[normalem]{ulem}
\usepackage{subfigure}
\usepackage{bm}

\begin{document}
%
\title{Object Detection based Deep Unsupervised Hashing}
\author{Rong-Cheng Tu, Xian-Ling Mao, Bo-Si Feng, Bing-Bing Bian, Yu-shu Ying\\
	School of Computer Science, Beijing Institute of Technology
}
\maketitle
\section{Abstract}
Recently, similarity-preserving hashing methods have been extensively studied for large-scale image retrieval. Compared with unsupervised hashing, supervised hashing methods for labeled data have usually better performance by utilizing semantic label information. Intuitively, for unlabeled data, it will improve the performance of unsupervised hashing methods if we can first mine some supervised semantic 'label information' from unlabeled data and then incorporate the 'label information' into the training process. Thus, in this paper, we propose a novel Object Detection based Deep Unsupervised Hashing method (ODDUH). Specifically, a pre-trained object detection model is utilized to mining supervised 'label information', which is used to guide the learning process to generate high-quality hash codes.Extensive experiments on two public datasets demonstrate that the proposed method outperforms the state-of-the-art unsupervised hashing methods in the image retrieval task. 
\section{Introduction}
\noindent With the rapid growth of image data, approximate nearest neighbour (ANN) search have attracted more and more attention from researchers in the large scale image search area. Among the existing ANN search techniques, similarity-preserving hashing methods are advantageous due to their high retrieval efficiency and low storage cost. The main idea of hashing methods are to transform high dimensional data points into a set of compact binary codes, meanwhile, maintain similarity of the original data points. Since the original data points are represented by binary codes instead of real valued features, the time and memory cost of searching can be dramatically reduced.

In general, data-dependent hashing can be divided into unsupervised \cite{gong2011iterative,huang2017unsupervised,ghasedi2018unsupervised,lin2016learning} and supervised \cite{zhu2016deep,qiu2017deep,zhang2018instance} methods. The unsupervised hashing methods mainly utilize the features of images to generate similarity-preserving binary codes without any supervised imformation. Compared with unsupervised hashing, supervised hashing methods incorporate sematic labels of training data into training process, thus they can perform more remarkably in generating similarity-preserving binary code. However, in many real applications, there are no semantic labels of images that can be used as supervised information. Hence, we just can use the unsupervised hashing methods to tackle the large scale image retrieval task in these case. 
\begin{figure}[]
	\centering
	\includegraphics[width=\linewidth]{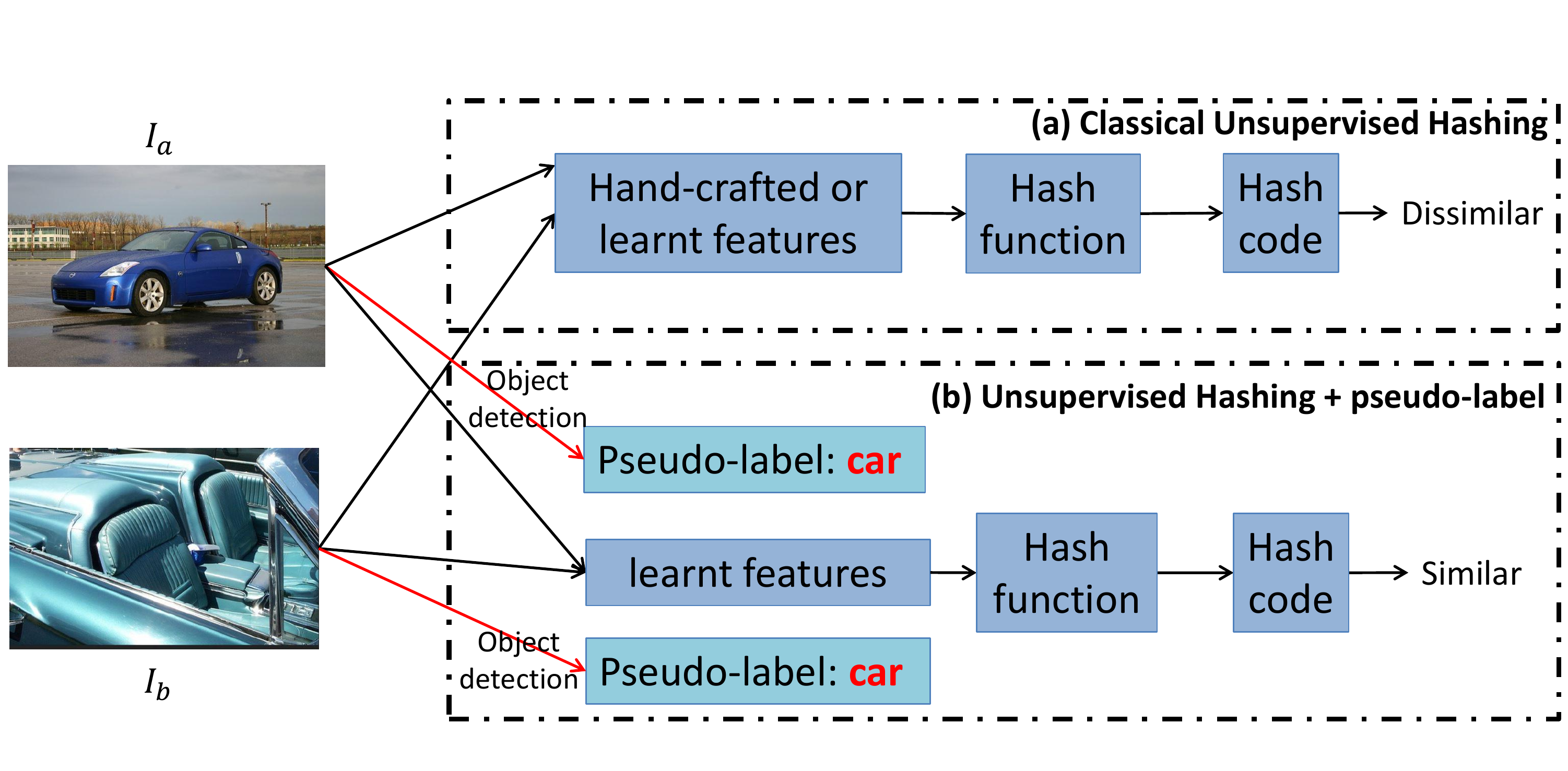}
	\caption{High-quality similarity-preserving hashing code can be produced by utilizing the pseudo-labels mined from images. The block (a) is the workflow of the existing unsupervised hashing methods which do not mine the pseudo-labels from images, it is hard for them to judge that the two images $I_a$ and $I_b$ are similar. However, in block (b), we use pseudo-labels mined from images to train hashing models which can easily judge the two image are similar.}
	\label{fig_illustrator}
\end{figure}

Intuitively, if we can detect the objects in images and use their classes as the pseudo-labels of the images, then we can use the  pseudo-labels as 'supervised information' to guide hash codes learning to obtain batter performance. An illustrative example is shown in Figure \ref{fig_illustrator}, the block (a) is the procedure of existing unsupervised hashing methods. They use the hand-crafted or learnt features as inputs. And they directly use the Euclidean distance between images or the similarity between one image and its rotated image to guide the hash training, which will make the existing unsupervised hashing models to judge that the images $I_a$ and $I_b$ are dissimilar with high possibility. Actually, the images $I_a$ and $I_b$ are similar, that the two images are belong to the class '$car$'. On the contrary, in the block (b), if we use an object detection model to get the pseudo-labels of the two images by detecting objects inside an image and classify each object into one of many different classes. And by utilizing the pseudo-labels that both classes of images are '$car$', we can construct pair-wise similarity to train hashing models and make the hashing models to judge that the two images $I_a$ and $I_b$ are similar with high possibility.

Inspired by this idea, we propose a novel Object Detection based Deep Unsupervised Hashing model, called ODDUH. In particular, an object detection model is first pre-trained on a large database which contains all the tags belonging to the hashing dataset. Then, we utilize the object detection model to mine latent semantic 'label information'( i.e., pseudo-labels) from images. And by taking use of pseudo-labels learned from the pre-trained object detection model, we define a novel similarity criterion called pair-wise percentage similarity inspired by ISDH \cite{zhang2018instance} . Moreover, a shared CNN is introduced to capture the feature representations of images. Finally, we combine the pair-wise percentage similarity and the learnt feature representations of images to learn hash functions and generate high-quality similarity-preserving hash codes.

Extensive experiments on two real-world public datasets illustrate that our method outperforms the state-of-the-art unsupervised hash methods in image retrieval tasks. Our main contributions are outlined as follows:
\begin{itemize}
\item We propose a novel unsupervised hashing architecture by introducing an pre-trained object detection model to mining semantic 'labels information' from images.
\item Pair-wise percentage similarity is the first used in unsupervised hashing methods. With the guiding of  pair-wise percentage similarity, we can greatly take use of the power of deep models to learn high-quality similarity-preserving binary codes. Moreover, the binary codes can preserve ranking imformation.
\item Experiments have shown that the proposed method can perform batter than the existing unsupervised hashing methods in large-scale image retrieval tasks.
\end{itemize}
\section{2.Related Work}
\subsection{Similarity-preserving Hashing}
Generally, existing hashing methods can be divided into data-independent hashing and data-dependent hashing. For data-independent hashing methods, the hashing functions are typically randomly generated without any training data. The representative data-independent methods include Locality Sensitivity Hashing (LSH) \cite{gionis1999similarity} and  its variants \cite{datar2004locality,kulis2009fast}. For data-dependent hashing methods, they can achieve better accuracy with shorter codes by learning hash functions from training data. Futhermore, data-dependent can be further classified into two categories: supervised \cite{wang2017supervised,liu2012supervised,song2015top} and unsupervised \cite{gionis1999similarity,jin2014density,ghasedi2018unsupervised} methods. The supervised methods can achieve remarkable performance by utilizing labeled data to learn hashing functions. And the label information in supervised hashing methods can be used in the following three ways: point-wise label, pair-wise labels and triplet. Representative point-wise label based methods include: supervised discrete hashing (SDH) \cite{shen2015supervised}. Representative pair-wise labels based deep hashing methods include: Deep Supervised Hashing with Pairwise (DPSH) \cite{li2015feature}, Deep Supervised Discrete Hashing (DSDH) \cite{li2017deep}, Supervised Hierarchical Deep Hashing (SHDH) \cite{wang2017supervised}. Representative triplet based deep hashing methods include: Deep Semantic-preserving and Ranking-based Hashing (DSRH) \cite{yao2016deep}, Deep Semantic Hashing with GANs (DSH-GANs) \cite{qiu2017deep}.

The unsupervised hashing can be divided into traditional unsupervised hashing methods and deep unsupervised hashing methods. The traditional unsupervised hashing methods used hand-crafted features and shallow hash functions to obtain binary hash code. Lots of algorithms in this category have been proposed, including Spectral Hashing (SH) \cite{weiss2009spectral}, Iterative Quantization (ITQ) \cite{gong2011iterative}. However, limited by the hand-crafted features and shallow hash functions, it is hard for them to deal with complex and high dimensional real-world data and keep the semantic similarity between original data in the binary hash codes. The deep unsupervised hashing methods utilize deep architecture to learn hash code. Among the deep unsupervised hashing methods, Deepbit \cite{lin2016learning} get rotation invariant and balanced binary hash codes by defined a quantization loss. Unsupervised triplet hashing (UTH) \cite{huang2017unsupervised} employs an unsupervised triplet loss to get  balanced hash codes. HashGAN \cite{ghasedi2018unsupervised} generate compact hash codes by a generative adversarial hashing network.

However, few existing unsupervised hashing methods take a good use of the latent sematic 'label information' in the images. Thus, in this paper, we propose a novel deep unsupervised hashing model based on object detection to generate high-quality hash codes by mining the latent semantic 'label information' contained in images and incorporating the 'label information' into the training process.
\subsection{Object Detection} 
Object detection has been studied widely for locating an object inside the image and classifying the object into one of many different categories. And object detection can be simple categorized into two categories: classical models and deep learning models. In the classical models, one of the most popular is Viola-Jones framework \cite{jones2001robust}, and it works by generating different (possibly thousands) simple binary classifiers using Haar features. However, with the growing success of deep learning, deep learning models are now state of the art in object detection, and many studies have been published about deep object detection. In this category, some models are based on region proposal which can solve the sliding windows problem, such as R-CNN \cite{girshick2014rich}; and some models are based on regression which can have a fast detection speed, e.g., YOLO \cite{redmon2016you}, SSD \cite{liu2016ssd}.

\begin{figure*}[tb]
	\centering
	\includegraphics[width=\textwidth]{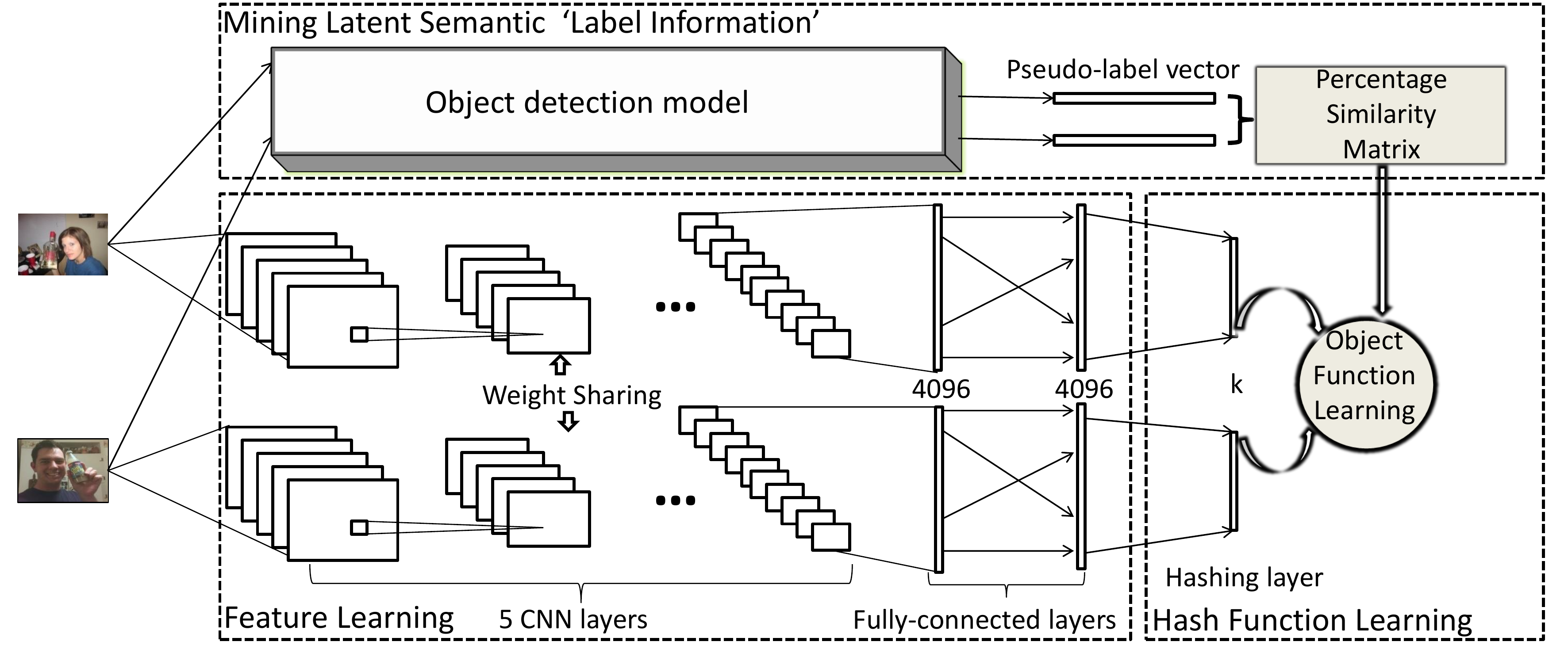}
	\caption{The ODDUH learning framework. The Mining Latent Semantic 'Label Information' is a pre-trained object detection model. It is used to get the pseudo-labels. A shared CNN is implemented for learning image fearture representations in the Feature Learning part. In Hash Function Learning part, a pair-wise loss function with Percentage Similarity Matrix is minimized to get the optimal hash function}
	\label{fig_architecture}
\end{figure*}
Most of the above approaches can have a good detection effect, and can mine the latent semantic 'label information' in images, which is exactly what our hashing architecture need. Thus, we can chose one of state-of-the-art object detection methods such as YOLOv2 \cite{redmon2017yolo9000} as a part of our hashing architecture.
\section{Object Detection based Deep Unsupervised Hashing Network}
In this section, we will present the proposed Object Detection based Deep Unsupervised Hashing Network (ODDUH) in detail.
\subsection{Notation}
Suppose a dataset has n images $X = \{x_i\}_{i=1} ^n$, and the $i^{th}$ image is $x_i$. The goal of similarity-preserving hashing is to learn a mapping $H:x_i \rightarrow \textbf b_i \in \{-1,1\}^k$, where $k$ is the length of hashing codes, such that an input image $x_i$ will be encoded into a k-bit binary code $\textbf b_i $.
\subsection{The Architecture of ODDUH}
As shown in Figure \ref{fig_architecture}, our architecture consists of three parts: mining latent semantic 'label information', feature learning and hash function learning. 

In the  mining latent semantic 'label information' part, the ODDUH uses a pre-trained object detection model named YOLOv2 \cite{redmon2017yolo9000} to mining the latent semantic 'label information' in images. Note that other state-of-the-art object detection models can also be used such as SSD \cite{liu2016ssd} and Mask R-CNN \cite{he2017mask}

The feature learning part includes a convolutional neural network (CNN) component which has five convolutional layers and two fully-connected layers. What's more, all the seven layers are the same as those of CNN-F network in Alexnet \cite{krizhevsky2012imagenet}.Note that other CNN architectures can also be used here, such as VGG \cite{simonyan2014very} and GoogLeNet \cite{szegedy2015going}.

The hash function learning part is a hashing layer which has $k$ units. Furthermore, $k$ is the length of hash code, and the hashing functions are learnt by the hashing layer. Eventually, we use element-wise sign function $sgn(\cdot)$, which returns $1$ if the element is positive and returns $-1$ otherwise, to process the outputs of the hashing layer and get the binary code $\textbf b$.
\subsection{Similarity Definition}
In ODDUH, we use an object detection model to mine the objects in images and get their classes (i.e., pseudo-labels). And for many images, more than one pesudo-labels will be mined. Thus, the unlabeled training dataset will become a 'mutil-label' dataset. In oder to take a good use of the mined semantic 'lable information', inspired  by ISDH \cite{zhang2018instance}, the pair-wise percentage similarity is defined as:
\begin{equation}
s_{ij}=\frac{\langle l_i,l_j \rangle}{\|l_i\|_2\|l_j\|_2} \label{1}
\end{equation}
where $\langle l_i,l_j \rangle$ calculate the inner product and $l_i \in \{0, 1\}^c$ is the pseudo-label vector of $x_i$, where c is the total number of classes that pseudo-labels belong to. If $i^{th}$ image $x_i$ has the $j^{th}$ pseudo-label, then $l_{ij} = 1$, else $l_{ij}=0$.

By incorporating the pair-wise percentage similarity into the training process, the learnt binary codes $\textbf B = \{ \textbf b_i\}_{i=1}^n$ can preserve the similarity in  $S = \{s_{ij}|i,j\in\{1,2,\dots,n\}, s_{ij}\in[0,1]\}$. More specifically, if $s_{ij}=0$, the binary codes $\textbf b_i$ and $\textbf b_j$ should have large Hamming distance; if $s_{ij}=1$, the binary codes $\textbf b_i$ and $\textbf b_j$ should have a small Hamming distance; otherwise, the binary codes $\textbf b_i$ and $\textbf b_j$ should have a suitable Hamming distance complying with the similarity $s_{ij}$.
\subsection{Objective Fuction}
Given the binary codes $\textbf B = \{\textbf b_i\}_{i=1}^n$ for all the images, we can define the likelihood of the pair-wise percentage similarity $s_{ij}$ as:
\begin{equation}
p(s_{ij}|\textbf B) = \left\{
\begin{array}{lrc}
\sigma(\Psi_{ij}),       && {s_{ij} = 1},\\
1 - \sigma(\Psi_{ij}),     && {s_{ij} = 0},\\
1- (s{ij} - \sigma(\Psi_{ij})),    && {otherwise}.
\end{array} \right. \label{2}
\end{equation}
where $\Psi_{ij}=\frac{1}{2}\textbf b_i^T\textbf b_j$, and $\sigma(\Psi_{ij}) = \frac{1}{1+e^{-\Psi_{ij}}}$. When $s_{ij} = 0\ or\ 1$, we take the negative log-likelihood of the observed pair-wise labels in S  to measure the pair-wise similarity loss , and when $0<s_{ij}<1$, we take  mean square error to measure the pair-wise similarity loss. Thus, the pair-wise similarity loss function can be defined as:
\begin{equation}
\begin{split}
L_1=&\sum_{s_{ij} \in S}[-\alpha\cdot I_{ij}(s_{ij}log(\sigma(\Psi_{ij})) \\
&+ (1-s_{ij})log(1-\sigma(\Psi_{ij})))\\
&+ (1 - I_{ij})(s_{ij}-\sigma(\Psi_{ij}))^2]\\
= &\sum_{s_{ij} \in S}[\alpha\cdot I_{ij}(log(1 + e^{\Psi_{ij}}) - s_{ij}\Psi_{ij})\\
&+ (1 - I_{ij})(s_{ij}-\sigma(\Psi_{ij}))^2] \label{3} 
\end{split}
\end{equation}
where $\alpha$ is a hyper-parameter. $I_{ij}$ is used to denote two conditions, $I_{ij} = 1$ when the pseudo-labels of $i^{th}$ image and the pseudo-labels of $j^{th}$ image are completely similar or dissimilar, i.e., $s_{ij} = 1\ or\  0$, and $I_{ij} = 0$ when the pseudo-labels of $i^{th}$ image and the pseudo-labels of $j^{th}$ image are partly similar, i.e., $0 < s_{ij} < 1$.

By minimizing Eq. (\ref{3}), we can make the hamming distance between two completely similar points as small as possible, and simultaneously make the hamming distance between two dissimilar points as large as possible. Meanwhile, we can make the partly similar image $x_i$ and image $x_j$ have the suitable hamming distance complying with the similarity $s_{ij}$.

However, Eq. (\ref{3}) is a discrete optimization problem, which is difficult to solve. Following previous work \cite{li2015feature}, we reformulated Eq. (\ref{3}) as:
\begin{equation}
\begin{aligned}
L_2 = &\sum_{s_{ij} \in S}[\alpha\cdot I_{ij}(log(1 + e^{\Theta_{ij}}) - s_{ij}\Theta_{ij})\\
&+ (1 - I_{ij})(s_{ij}-\sigma(\Theta_{ij}))^2]
\end{aligned}
\end{equation}
where $\Theta_{ij} = \frac 12 \textbf u_i^T \textbf u_j$. $\textbf u_i \in \mathbb{R}^k$ is the outputs of hashing layer: $\textbf u_i = \textbf W^T \mathcal{F}(\textbf x_i;\bm{\theta}) + \textbf v$, where the mapping $\mathcal{F}: \mathbb{R}^d \rightarrow \mathbb{R}^{4096}$ is parameterized by $\bm{\theta}$ and $\bm{\theta}$ represents the parameters of the seven layers of network in the feature learning part. $\textbf W \in \mathbb{R}^{4096 \times k}$ is the weight matrix to be learnt at the hashing layer, $\textbf v \in \mathbb{R}^k$ is the bias.
Due to the $\textbf u_i$ is not the binary codes, we used a quantization loss to make $\textbf u_i$ to be close to binary codes. The quantization loss is defined as:
\begin{equation}
L_q = \sum_i^n||\textbf b_i - \textbf u_i||_2^2
\end{equation}
Then, by connecting the pseudo-label pair-wise similarity loss and quantization loss, the final objective function can be defined as:
\begin{equation}
\begin{aligned}
L = L_2 + \beta L_q
\end{aligned}
\end{equation}
where $\beta$ is a hyper-parameter.
\subsection{Learning}
In our method, the parameters containing $\textbf B, \textbf W,\bm{\theta},\textbf v$ need to be learnt, during the training phase. A  mini-batch gradient descent method is used for learning. Moreover, we design an alternating method for learning. More specifically, we optimize $\textbf B$ with $\textbf W,\bm{\theta},\textbf v$ fixed and optimize  $\textbf W,\bm{\theta},\textbf v$ with $\textbf B$ fixed.

The $\textbf b_i$ can be directly optimized as follows:
\begin{equation}
\textbf b_i = sgn(\textbf u_i) = sgn(\textbf W^T \mathcal{F}(\textbf x_i;\bm{\theta}) + \textbf v)
\label{b}
\end{equation}

For the other parameters $\textbf W,\bm{\theta},\textbf v$, standard back-propagation algorithm is used for learning. Especially, we are able to compute the derivatives of the loss function about $\textbf u_i$ as follows:
\begin{equation}
\begin{aligned}
\frac{\partial L}{\partial \textbf u_i} = &\sum_{j:s_{ij} \in S}[ \frac{1}{2} \alpha \cdot I_{ij} (\sigma(\Theta_{ij}) - s_{ij})  \\
&+ (1 - I_{ij})\sigma(\Theta_{ij})(1-\sigma(\Theta_{ij}))(s_{ij}-\sigma(\Theta_{ij}))]\textbf u_j	\\
&+ \sum_{j:s_{ji} \in S} [\frac{1}{2} \alpha \cdot I_{ji} (\sigma(\Theta_{ji}) - s_{ji}) \\
&+ (1 - I_{ji})\sigma(\Theta_{ji})(1-\sigma(\Theta_{ji}))(s_{ji}-\sigma(\Theta_{ji}))]\textbf u_j \\
&+2\beta(\textbf u_i - \textbf b_i) \label{7}
\end{aligned}
\end{equation}
Then, we can use the standard back-propagation algorithm to update $\textbf W,\bm{\theta}$ and $\textbf v$ with Eq. (\ref{7}):
\begin{equation}
\frac{\partial L}{\partial \textbf W} = \mathcal{F}(\textbf x_i;\bm{\theta}) (\frac{\partial L}{\partial \textbf u_i})^T
\label{w}
\end{equation}
\begin{equation}
\frac{\partial L}{\partial \mathcal{F}(\textbf x_i;\bm{\theta})} = \textbf W \frac{\partial L}{\partial \textbf u_i}
\label{s}
\end{equation}
\begin{equation}
\frac{\partial L}{\partial \textbf v} = \frac{\partial L}{\partial \textbf u_i}
\label{v}
\end{equation}
The outline of the proposed method is described in Algorithm 1.

\begin{algorithm}
	\caption{Learning algorithm for ODDUH}
	\label{alg:A}
	\begin{algorithmic}[1]
		\REQUIRE
		Training images $X = \{x_i\}_{i=1} ^n$, the max iterative count E, the size of mini-batch(default 128), the length of  hash codes K.
		\ENSURE 
		The hash codes for of images.
		\STATE Initialize the weights and bias of the Hashing model.
		\STATE Initialize all the hyper-parameters$\alpha, \beta$ and learning rate $r$ as 2, 100, 0.01 respectively.
		\STATE Utilize pre-trained object detection model to get pseudo label vectors of all the images.
		\STATE $S \leftarrow$ using Eq. (\ref{1}), $S \in \mathbb{R}^{N\times N}$
		\REPEAT 
		\STATE Update $r \leftarrow r/10$ every 50 iterations empirically.
		\STATE Randomly sample a mini-batch of images from $X$, and for each image $\textbf x_i$ , perform as follows:
		\STATE Calculate $\textbf u_i = \textbf W^T \mathcal{F}(\textbf x_i;\bm{\theta}) + \textbf v$;
		\STATE Calculate the hash code with Eq. (\ref{b});
		\STATE Update the parameters $\{\textbf W,\bm{\theta},\textbf v\}$ by back propagation with Eq. (\ref{w}), Eq. (\ref{s}) and Eq. (\ref{v}), respectively.
		\UNTIL Up to E
		\STATE Use Eq. (\ref{b}) calculate the hash codes of all the images. 
	\end{algorithmic}
\end{algorithm}
\section{Experiments}

\subsection{Dataset and Baseline}
We conduct experiments on two public benchmark datasets: Pascal VOC 2007 \footnote{\url{http://host.robots.ox.ac.uk/pascal/VOC/voc2007/}} \cite{everingham2010pascal} and BMVC 2009 \footnote{\url{http://pascal.inrialpes.fr/data2/flickr-bmvc2009/}} \cite{allan2009ranking}. Pascal VOC 2007 consists of 9,963 multi-label images. There are 20 object classes in this dataset. On average, each image is annotated with 1.5 labels. BMVC 2009 contains 96,378 images collected from Flickr. Each image in the dataset is associated with one or multiple labels in 20 semantic concepts.

Our proposed method is an unsupervised method, thus we compare our method with eight calssical and state-of-the-art unsupervised hashing methods including: 
LSH \cite{gionis1999similarity}, ITQ \cite{gong2011iterative}, SH \cite{weiss2009spectral}, PCAH \cite{wang2010semi}, SGH \cite{jiang2015scalable}, UH\_BDNN \cite{do2016learning}, UTH \cite{huang2017unsupervised}, HashGAN \cite{ghasedi2018unsupervised}, where LSH, SH, ITQ, PCAH and SGH are traditional unsupervised methods and the other three are deep unsupervised methods. Note that the five traditional unsupervised hashing methods use hand-crafted features as inputs. And each image in Pascal VOC 2007 and BMVC 2009 is represented by a 512-dimensional GIST vector. For the deep unsupervised hashing method UH\_BDNN, it use the outputs of fc7 layer in AlexNet as image representation. And for the other two deep unsupervised hashing methods and our proposed method, we resize all the images to be $224 \times 224$ pixels and then directly use the raw image pixels as input. When carry out experiments on the two datasets respectively, we randomly select 2,000 images as test set and the left images as training dataset. Moreover, we also conduct the experiments by using the outputs of fc7 layer in AlexNet  \cite{krizhevsky2012imagenet} as image representation in the five traditional hashing approaches and denote them as LSH+CNN, SH+CNN, ITQ+CNN and PCAH+CNN, respectively.
\subsection{Implementation details}
For the object detection component, we choose YOLOv2 \cite{redmon2017yolo9000}. And it is pre-trained in COCO 2014 dataset which contains 81 object classes. Please note that all the object classes contained in Pascal VOC 2007 and BMVC 2009 are subdet of the 81 object classes. For the hash code learning model formed by the feature learning part and hash function learning part, all the weights and bias are learned via back-propagation algorithm. Furthermore, the weights and bias in the feature learning part are initialized as the values pre-trained in Alexnet  \cite{krizhevsky2012imagenet}. We adopt SGD with a mini-batch size of 128 as our optimization algorithm. The learning rate is initialized as 0.01. The hyper-parameters $\alpha, \beta$ in ODDUH are empirically set as 2 and 100, respectively. We will discuss the effect of $\alpha, \beta$ in the followed subsection. And the learning rate is adjusted to one tenth of the current learning rate every one third of epoches.
\subsection{Evaluation criterion}
To verify the effectiveness of hash code, we evaluate the image retrieval quality for different methods by  Average Cumulative Gains (ACG), Normalized Discounted Cumulative Gains (NDCG), Mean Average Precision (MAP) and Weighted Mean Average Precision (W-MAP).

ACG@n represents the average of similarities between the query image and the top n retrieved images, which can be calculated as:
\begin{equation}
ACG@n = \sum_{j=1}^n \frac{r(j)}{n}
\end{equation}
where $r(j)$ is defined as the number of shared labels between the query image and the $j^{th}$ retrieved image.

NDCG is a widely used evaluation metric in information retrieval. Given a query image, the DCG score of top n retrieved images is defined as:
\begin{equation}
DCG@n = \sum_{j=1}^n \frac{2^{r(j)} - 1}{log(i+1)}
\end{equation}
Then, the normalized DCG (NDCG) score at the position n can be calculated by$ NDCG@n = \frac{DCG@n}{Z_n}$, where $Z_n$ is the maximum value of DCG@n, constraining the value of NDCG in the range [0,1].

MAP, a standard evaluation metric for information retrieval, is the mean of average precision for each query. It is defined as:
\begin{equation}
MAP = \sum_{j=1}^n P_j \frac{p(j)}{N}
\end{equation}
where $P_j = \frac{R(j)}{j}$, R(j)  represents the number of relevant images within the top j images. $p(j)$ is a indicator function, if the image at the position j shares at least one label with the query image, $p(j)$ is 1; otherwise $p(j)$ is 0. N represents the total number of relevant images, i.e., it shares at least one label with the query image.

The definition of W-MAP is similar with MAP, defined as:
\begin{equation}
WMAP = \sum_{j=1}^n ACG@j \frac{p(j)}{N}
\end{equation}

Furthermore, the three evaluation criterions WMAP, NDCG and ACG are usually used to measure the ranking quality of hashing models. Beacuse the greater value of WMAP or NDCG means that the related items in the retrieved result list have higher ranks. And the larger value of ACG indicates that the images in the retrieved result list are more similar to the query image. 
\begin{table*}[t]
	\begin{minipage}{\textwidth}
		\caption{Results on the Pascal VOC 2007. The ranking results are measured by NDCG, ACG, WMAP, and MAP@N (N=1000, i.e., the values are calculated based on the top 1000 returned neighbors).  The best results for each category are shown in boldface}
		\label{table_voc}
		\resizebox{\linewidth}{!}{
			\begin{tabular}{|c||c|c|c|c||c|c|c|c||c|c|c|c||c|c|c|c|}
				\hline
				\multirow{2}{*}{\textit{\textbf{Methods}}} & \multicolumn{4}{c||}{MAP@1000} & \multicolumn{4}{c||}{WMAP@1000} & \multicolumn{4}{c||}{NDCG@1000} & \multicolumn{4}{c|}{ACG@1000} \\ \cline{2-17} 
				& 12bits & 24bits & 36bits & 48bits & 12bits & 24bits & 36bits & 48bits & 12bits & 24bits & 36bits & 48bits & 12bits & 24bits & 36bits & 48bits \\ \hline \hline
				LSH & 0.2676 & 0.2875 & 0.2916 & 0.2877 & 0.2881 & 0.3115 & 0.3168 & 0.3123 & 0.2230 & 0.2379 & 0.2436 & 0.2407 & 0.2821 & 0.2998 & 0.3009 & 0.2968 \\ \hline 
				SH & 0.3071 & 0.3021 & 0.3028 & 0.3023 & 0.3337 & 0.3287 & 0.3299 & 0.3299 & 0.2568 & 0.2514 & 0.2527 & 0.2530 & 0.3131 & 0.3074 & 0.3071 & 0.3055 \\ \hline
				PCAH & 0.2884 & 0.2802 & 0.2783 & 0.2778 & 0.3124 & 0.3039 & 0.3018 & 0.3013 & 0.2384 & 0.2320 & 0.2307 & 0.23.5 & 0.2982 & 0.2883 & 0.2849 & 0.2837 \\ \hline
				ITQ & 0.2879 & 0.3086 & 0.3137 & 0.3223 & 0.3110 & 0.3345 & 0.3404 & 0.3509 & 0.2366 & 0.2584 & 0.2620 & 0.2718 & 0.2924 & 0.3191 & 0.3258 & 0.3358 \\ \hline
				SGH & 0.3028 & 0.3081 & 0.3073 & 0.3107 & 0.3288 & 0.3358 & 0.3350 & 0.3395 & 0.2559 & 0.2611 & 0.2614 & 0.2644 & 0.3052 & 0.3082 & 0.3065 & 0.3089 \\ \hline \hline
				LSH+CNN & 0.2924 & 0.3351 & 0.3611 & 0.3694 & 0.3226 & 0.3737 & 0.4001 & 0.4189 & 0.2502 & 0.2875 & 0.3030 & 0.3142 & 0.2993 & 0.3314 & 0.3505 & 0.3542 \\ \hline
				SH+CNN & 0.4497 & 0.4454 & 0.4585 & 0.4587 & 0.5122 & 0.5033 & 0.5162 & 0.5160 & 0.3927 & 0.3757 & 0.3780 & 0.3731 & 0.4065 & 0.3837 & 0.3853 & 0.3800 \\ \hline
				PCAH+CNN & 0.4892 & 0.4914 & 0.4890 & 0.4848 & 0.5515 & 0.5514 & 0.5486 & 0.5439 & 0.4337 & 0.4185 & 0.4066 & 0.3961 & 0.4454 & 0.4207 & 0.4067 & 0.3962 \\ \hline
				ITQ+CNN & 0.5606 & 0.5886 & 0.6006 & 0.6070 & 0.6429 & 0.6777 & 0.6927 & 0.6996 & 0.5137 & 0.5266 & 0.5323 & 0.5368 & 0.5328 & 0.5362 & 0.5366 & 0.5391 \\ \hline
				SGH+CNN & 0.2575 & 0.2653 & 0.2730 & 0.2839 & 0.2773 & 0.2871 & 0.2955 & 0.3083 & 0.2129 & 0.2198 & 0.2254 & 0.2339 & 0.2675 & 0.2718 & 0.2748 & 0.2789 \\ \hline \hline
				UH\_BDNN & 0.5572 & 0.5795 & 0.5851 & 0.5915 & 0.6388 & 0.6639 & 0.6700 & 0.6781 & 0.5080 & 0.5132 & 0.5110 & 0.5115 & 0.5188 & 0.5168 & 0.5101 & 0.5067 \\ \hline 
				UTH & 0.5389 & 0.5468 & 0.5561 & 0.5634 & 0.6192 & 0.6286 & 0.6427 & 0.6451 & 0.4856 & 0.4921 & 0.4994 & 0.5012 & 0.4961 & 0.4979 & 0.5006 & 0.5013 \\ \hline
				HashGAN & 0.4606 & 0.4672 & 0.4711 & 0.4783 & 0.5114 & 0.5201 & 0.5263 & 0.5310 & 0.4115 & 0.4183 & 0.4214 & 0.4240 & 0.4197 & 0.4246 & 0.4293 & 0.4303 \\ \hline
				ODUDH & \textbf{0.6946} & \textbf{0.7335} & \textbf{0.7538} & \textbf{0.7604} & \textbf{0.7449} & \textbf{0.7871} & \textbf{0.8083} & \textbf{0.8321} & \textbf{0.5979} & \textbf{0.6162} & \textbf{0.6249} & \textbf{0.6441} & \textbf{0.6206} & \textbf{0.6284} & \textbf{0.6324} & \textbf{0.6512} \\ \hline
		\end{tabular}}
	\end{minipage}
	\begin{minipage}{\textwidth}
		\caption{Results on the BMVC 2009. The ranking results are measured by NDCG, ACG, WMAP, and MAP@N (N=5000, i.e., the values are calculated based on the top 5000 returned neighbors).  The best results for each category are shown in boldface}
		\label{table_bmvc}
		\resizebox{\linewidth}{!}{
			\begin{tabular}{|c||c|c|c|c||c|c|c|c||c|c|c|c||c|c|c|c|}
				\hline
				\multirow{2}{3cm}{\textit{\textbf{Methods}}} & \multicolumn{4}{c||}{MAP@5000} & \multicolumn{4}{c||}{WMAP@5000} & \multicolumn{4}{c||}{NDCG@5000} & \multicolumn{4}{c|}{ACG@5000} \\ \cline{2-17} 
				& 12bits & 24bits & 36bits & 48bits & 12bits & 24bits & 36bits & 48bits & 12bits & 24bits & 36bits & 48bits & 12bits & 24bits & 36bits & 48bits \\ \hline \hline
				LSH & 0.1393 & 0.1494 & 0.1539 & 0.1494 & 0.1495 & 0.1602 & 0.1652 & 0.1604 & 0.1070 & 0.1136 & 0.1174 & 0.1143 & 0.1459 & 0.1543 & 0.1582 & 0.1536 \\ \hline
				SH & 0.1656 & 0.1629 & 0.1641 & 0.1668 & 0.1785 & 0.1756 & 0.1768 & 0.1796 & 0.1247 & 0.1221 & 0.1235 & 0.1252 & 0.1706 & 0.1664 & 0.1677 & 0.1694 \\ \hline
				PCAH & 0.1452 & 0.1464 & 0.1477 & 0.1499 & 0.1562 & 0.1575 & 0.1588 & 0.1614 & 0.1108 & 0.1110 & 0.1116 & 0.1125 & 0.1513 & 0.1516 & 0.1517 & 0.1533 \\ \hline
				ITQ & 0.1356 & 0.1423 & 0.1599 & 0.1618 & 0.1431 & 0.1508 & 0.1719 & 0.1738 & 0.0996 & 0.1081 & 0.1206 & 0.1237 & 0.1326 & 0.1439 & 0.1608 & 0.1656 \\ \hline 
				SGH & 0.1681 & 0.1698 & 0.1715 & 0.1724 & 0.1807 & 0.1825 & 0.1841 & 0.1850 & 0.1280 & 0.1287 & 0.1298 & 0.1305 & 0.1696 & 0.1700 & 0.1700 & 0.1701 \\ \hline \hline
				LSH+CNN & 0.1621 & 0.1925 & 0.1954 & 0.2133 & 0.1750 & 0.2078 & 0.2112 & 0.2304 & 0.1233 & 0.1425 & 0.1435 & 0.1544 & 0.1657 & 0.1887 & 0.1878 & 0.1976 \\ \hline 
				SH+CNN & 0.2667 & 0.2805 & 0.2798 & 0.2877 & 0.2880 & 0.3041 & 0.3032 & 0.3122 & 0.1845 & 0.1981 & 0.1931 & 0.1971 & 0.2452 & 0.2460 & 0.2379 & 0.2429 \\ \hline
				PCAH+CNN & 0.2991 & 0.3076 & 0.3063 & 0.3090 & 0.3236 & 0.3336 & 0.3321 & 0.3354 & 0.2215 & 0.2187 & 0.2122 & 0.2105 & 0.2803 & 0.2720 & 0.2616 & 0.2584 \\ \hline
				ITQ+CNN & 0.3330 & 0.3627 & 0.3712 & 0.3781 & 0.3617 & 0.3942 & 0.4034 & 0.4123 & 0.2527 & 0.2684 & 0.2726 & 0.2765 & 0.3223 & 0.3339 & 0.3370 & 0.3415 \\ \hline
				SGH+CNN & 0.1344 & 0.1423 & 0.1493 & 0.1575 & 0.1444 & 0.1530 & 0.1606 & 0.1697 & 0.1033 & 0.1079 & 0.1117 & 0.1152 & 0.1413 & 0.1460 & 0.1496 & 0.1527 \\ \hline \hline
				UH\_BDNN & 0.3442 & 0.3736 & 0.3828 & 0.3960 & 0.3737 & 0.4049 & 0.4148 & 0.4289 & 0.2605 & 0.2768 & 0.2811 & 0.2876 & 0.3262 & 0.3405 & 0.3439 & 0.3500 \\ \hline
				UTH & 0.3011 & 0.3083 & 0.3102 & 0.3138 & 0.3375 & 0.3417 & 0.3481 & 0.3495 & 0.2276 & 0.2292 & 0.2317 & 0.2343 & 0.2743 & 0.2835 & 0.2891 & 0.2931 \\ \hline
				HashGAN & 0.2711 & 0.2790 & 0.2866 & 0.2935 & 0.2930 & 0.3052 & 0.3121 & 0.3167 & 0.2028 & 0.2091 & 0.2131 & 0.2177 & 0.2496 & 0.2528 & 0.2589 & 0.2635 \\ \hline
				ODUDH & \textbf{0.4091} & \textbf{0.4228} & \textbf{0.4254} & \textbf{0.4313} & \textbf{0.4428} & \textbf{0.4570} & \textbf{0.4582} & \textbf{0.4651} & \textbf{0.3205} & \textbf{0.3259} & \textbf{0.3274} & \textbf{0.3306} & \textbf{0.4047} & \textbf{0.4085} & \textbf{0.4093} & \textbf{0.4130} \\ \hline
		\end{tabular}}
	\end{minipage}
\end{table*}
\begin{figure*}[t]
	\centering
	\subfigure[]{
		\begin{minipage}[t]{0.5\textwidth}
			\centering
			\includegraphics[width=\linewidth]{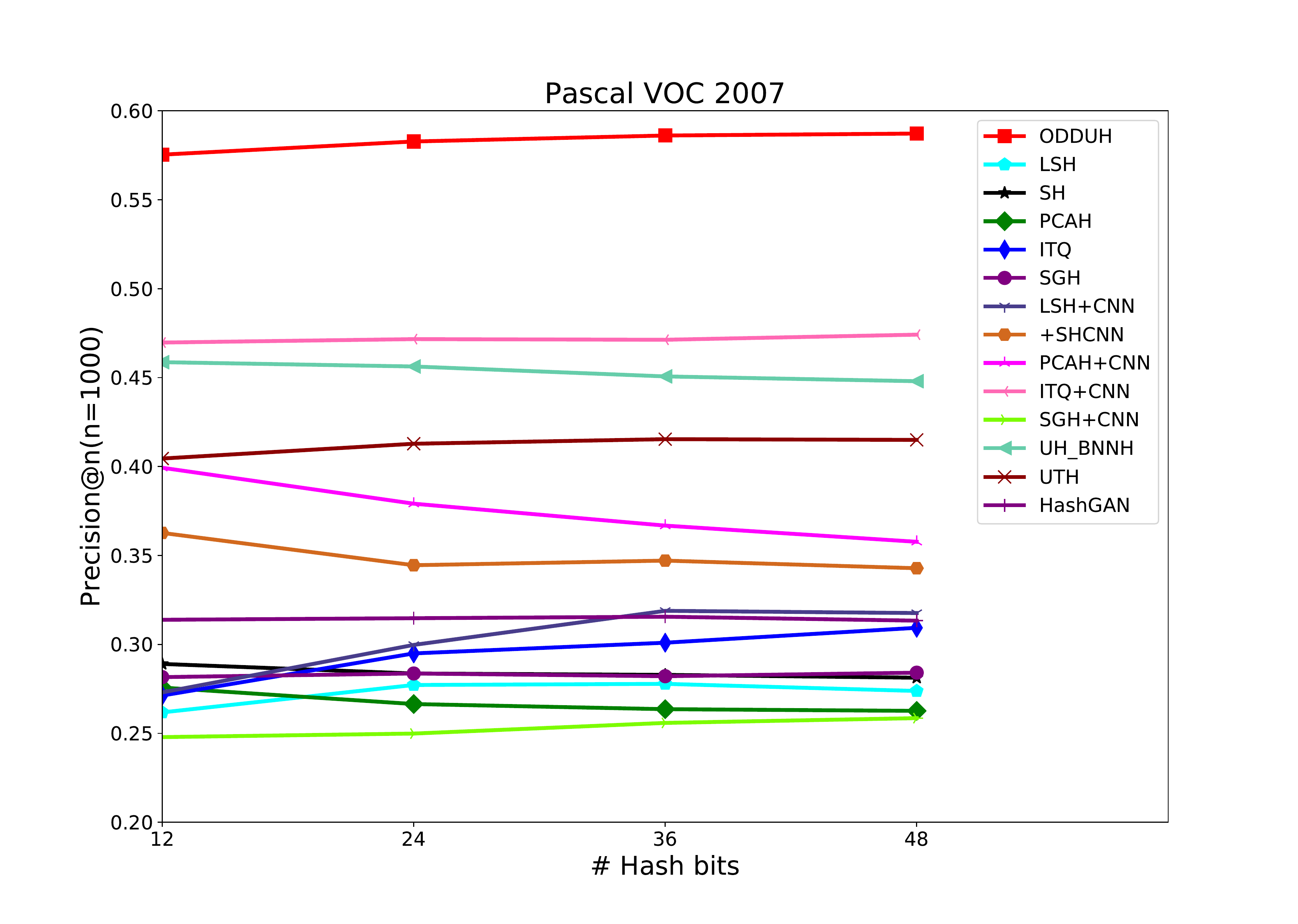}
		\end{minipage}%
	}%
	\subfigure[]{
		\begin{minipage}[t]{0.5\textwidth}
			\centering
			\includegraphics[width=\linewidth]{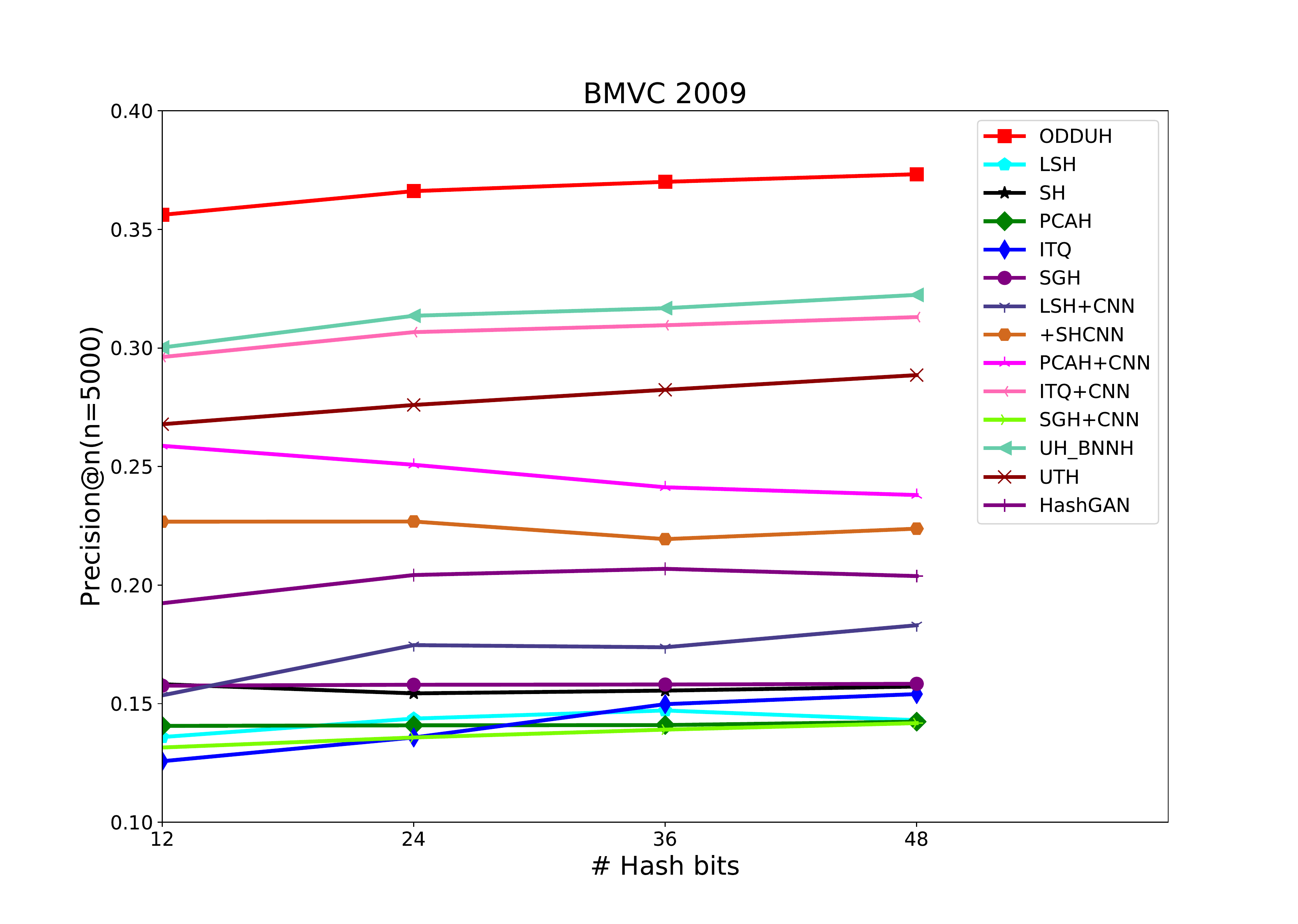}
		\end{minipage}%
	}%
	\caption{Precision@n over (a) Pascal VOC 2007 and (b) BMVC 2009}
	\label{fig_precision}
\end{figure*}
\begin{figure*}[]
	\centering
	\subfigure[Sensitivity to hyper-parameter $\alpha$ over two dataset with $\beta=100$]{
		\begin{minipage}[t]{0.5\textwidth}
			\centering
			\includegraphics[width=\linewidth]{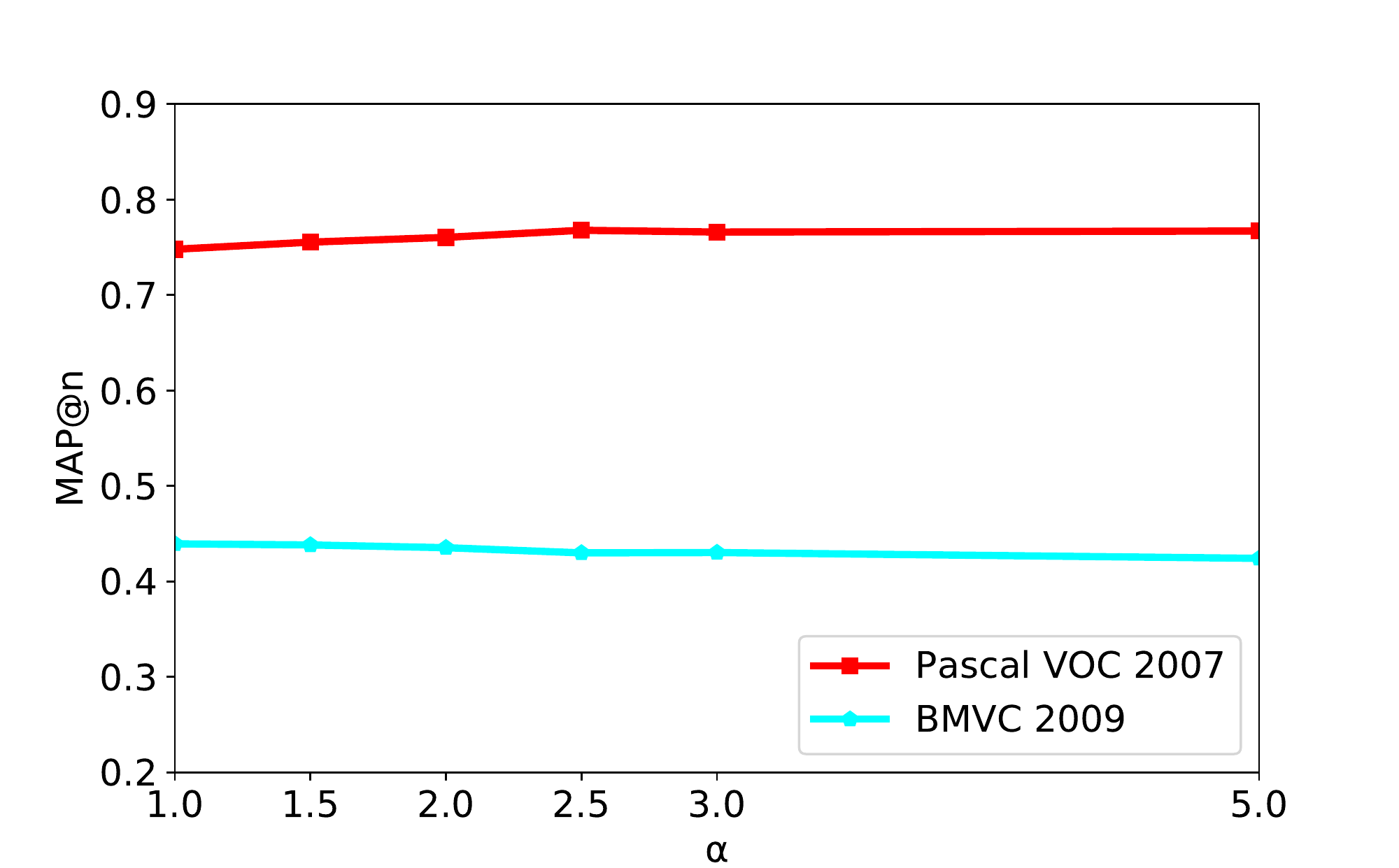}
			\label{fig_alpha}
		\end{minipage}%
	}%
	\subfigure[Sensitivity to hyper-parameter $\beta$ over Pascal VOC 2007 with $\alpha=2$]{
		\begin{minipage}[t]{0.5\textwidth}
			\centering
			\includegraphics[width=\linewidth]{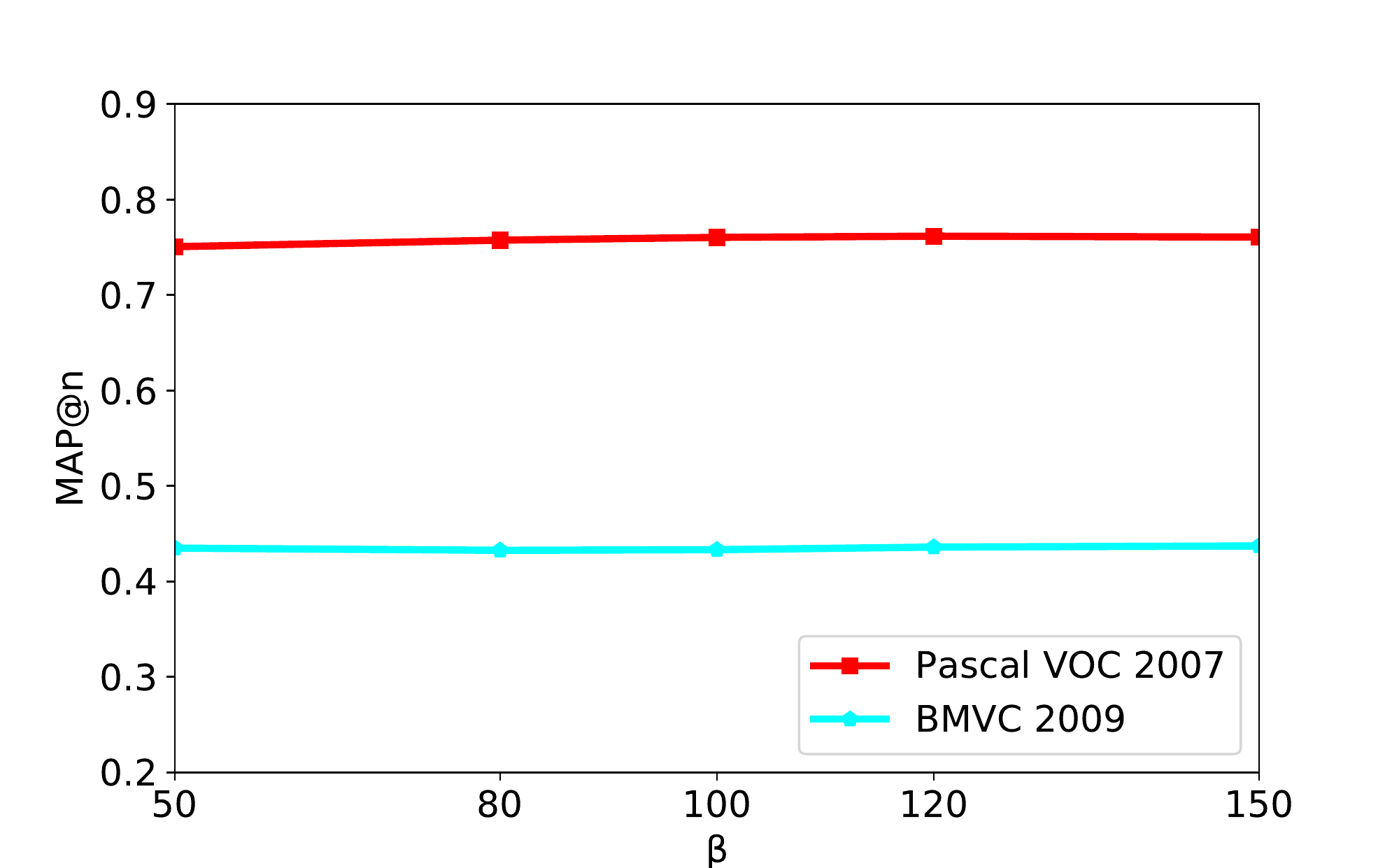}
			\label{fig_beta}
		\end{minipage}%
	}%
	\caption{ Sensitivity to hyper-parameters}
	\label{fig_params}
\end{figure*}
\subsection{Experimental results}
Table \ref{table_voc} summarizes the comparative results of different hashing methods over Pascal VOC 2007. And table \ref{table_bmvc}  shows the performance comparison of different hashing methods over BMVC 2009. In general, It can be found that our proposed method substantially outperforms the other unsupervised hashing methods for different length of hash code.  In particular, on Pascal VOC 2007, comparing with the best traditional competitor ITQ+CNN on 48-bits, the results of ODDUH have a relative increase of 25.3\% on MAP, 18.9\% on WMAP, 20.0\% on NDCG, 20.8\% on ACG, and comparing with the best deep unsupervised competitor UH\_BDNN on 48-bits, the results of ODDUH have a relative increase of 25.9\% on MAP, 22.7\% on WMAP, 25.9\% on NDCG, 28.5\% on ACG. Moreover, on BMVC 2009, comparing with the competitor UH\_BDNN on 48-bits, the results of ODDUH have a relative increase of 8.9\% on MAP, 8.4\% on WMAP, 15.0\% on NDCG, 18.0\% on ACG. And as the growth of the three evaluation criterions WMAP, NDCG, ACG illustrate that the ODDUH can preserve more ranking information in the binary codes than the baselines. Furthermore,the results obviously indicates that pseudo-labels mined from images are more advantageous than the Euclidean distance between images or the similarity between one image and its rotated images to generate similarity-preserving hashing codes. In addition, most traditional unsupervised methods with image representations extracted from deep CNN architecture perform better than these methods with GIST features, which proves that the learnt representations by deep network from raw images are more superior than hand-crafted features in hash learning procedure.

The precision@n ( the precision value is calculated based on the top n returned neighbors) curves at different length of hash code are also showed in Figure \ref{fig_precision}. Figure \ref{fig_precision}(a) and Figure \ref{fig_precision}(b) exhibit the precision@n curves over Pascal VOC 2007 and BMVC 2009, respectively. And both Figure \ref{fig_precision}(a) and Figure \ref{fig_precision}(b) show that our ODDUH model perform batter than baselines. Also, the precisions for most of the baselines drop when the length of hash codes increases. Contrarily, the precision of ODDUH is still growing, which means ODDUH is more stable.
\subsection{Sensitivity to Hyper-Parameters}
Figure \ref{fig_alpha} shows the effect of the hyper-parameter $\alpha$ on 48 bits over Pascal VOC 2007 and BMVC 2009. It can be found that  ODDUH is not sensitive to $\alpha$ on both datasets. For instance, ODDUH can achieve good performance on both datasets with $1 \leq \alpha \leq 5$. Figure \ref{fig_beta} shows the effect of the hyper-parameter $\beta$ on 48 bits over Pascal VOC 2007 and BMVC 2009. Also, ODDUH is not sensitive to $\beta$ in a large range. For example, ODDUH can achieve good performance on both datasets with $80 \leq \beta \leq 150$. And we can also obtain similar conclusion on other length of hash codes for both hyper-Parameters $\alpha$ and $\beta$.
\section{Conclusion}
In this paper, we have proposed a novel Object Detection based Deep Unsupervised Hashing method, called, for unlabeled data. To the best of our knowledge, ODDUH is the first method which utilize object detection model to mine semantic 'label information' from images. By  incorporating the semantic 'label information' into the training process, the learnt hashing functions can generat high-quality similarity-preserving hash codes. Extensive experiments on two real-world public datasets have shown that the proposed ODDUH method can outperform other methods to achieve the state-of-the-art performance in image retrieval applications.
\bibliographystyle{plain}
\bibliography{19}
\end{document}